\documentclass[letterpaper, 10 pt, conference]{ieeeconf}  % Comment this line out if you need a4paper
\usepackage{amsfonts}
\usepackage{booktabs} 
\usepackage{algorithm}
\usepackage{algorithmic}  
\usepackage{tabularx} 
\usepackage{multirow}
\usepackage{xcolor}
\usepackage{booktabs}
\usepackage{adjustbox}
\usepackage{xcolor}
\usepackage{multirow}
\usepackage{pifont}
\newcolumntype{Y}{>{\raggedright\arraybackslash}X}
\definecolor{deepgray}{HTML}{808080}
\definecolor{deepblue}{HTML}{1459a8}

\IEEEoverridecommandlockouts                              % This command is only needed if 
\usepackage{graphicx} 
 \usepackage{amsmath}
 \usepackage{hyperref}
\title{\LARGE \bf
DriveAgent: Multi-Agent Structured Reasoning with LLM and Multimodal Sensor Fusion for Autonomous Driving
}

\author{
Xinmeng Hou$^{2,*}$, Wuqi Wang$^{1,*}$, Long Yang$^{1}$, Hao Lin$^{3}$, Jinglun Feng$^{4,\dagger}$, Haigen Min$^{1,\dagger}$, Xiangmo Zhao$^{1}$%
\thanks{$^{1}$ Wuqi Wang, Long Yang, Haigen Min, and Xiangmo Zhao are with Chang'an University, Xi'an, Shaanxi, China.}%
\thanks{$^{2}$ Xinmeng Hou is with Chang'an University, Xi'an, Shaanxi, China and Agency for Science, Technology and Research (A*STAR), Singapore.}%
\thanks{$^{3}$ Hao Lin is with University of California, Davis, USA.}%
\thanks{$^{4}$ Jinglun Feng is with CCNY Robotics Lab, The City College of New York, USA. 
   {$^*$ Equally contributed.  $^\dagger$ Corresponding authors.}}
\thanks{Jinglun Feng:{jfeng1@ccny.cuny.edu},Haigen Min:{hgmin@@chd.edu.cn}}%
 \thanks{This work is supported in part by the National Natural Science Foundation (52441205), National Natural Science Foundation of China (No.52372426), Shaanxi Province Innovation Capability Support Plan-Innovative Talent Promotion Plan (No.2023KJXX-020), Natural Science Foundation of Shaanxi Province (No.2022JQ-663), Key Research and Development Program of Shaanxi Province (2024GX-YBXM-261), and Fundamental Research Funds for the Central Universities, CHD (No.300102243202, 300102244713).}
}

\begin{document}

\maketitle
\thispagestyle{empty}
\pagestyle{empty}

%%%%%%%%%%%%%%%%%%%%%%%%%%%%%%%%%%%%%%%%%%%%%%%%%%%%%%%%%%%%%%%%%%%%%%%%%%%%%%%%
\begin{abstract}
We introduce DriveAgent, a novel multi-agent autonomous driving framework that leverages large language model (LLM) reasoning combined with multimodal sensor fusion to enhance situational understanding and decision-making. DriveAgent uniquely integrates diverse sensor modalities—including camera, LiDAR, GPS, and IMU—with LLM-driven analytical processes structured across specialized agents. The framework operates through a modular agent-based pipeline comprising four principal modules: (i) a descriptive analysis agent identifying critical sensor data events based on filtered timestamps, (ii) dedicated vehicle-level analysis conducted by LiDAR and vision agents that collaboratively assess vehicle conditions and movements, (iii) environmental reasoning and causal analysis agents explaining contextual changes and their underlying mechanisms, and (iv) an urgency-aware decision-generation agent prioritizing insights and proposing timely maneuvers. This modular design empowers the LLM to effectively coordinate specialized perception and reasoning agents, delivering cohesive, interpretable insights into complex autonomous driving scenarios. Extensive experiments on challenging autonomous driving datasets demonstrate that DriveAgent is achieving superior performance on multiple metrics against baseline methods. These results validate the efficacy of the proposed LLM-driven multi-agent sensor fusion framework, underscoring its potential to substantially enhance the robustness and reliability of autonomous driving systems. \footnote{Code available at \url{https://github.com/Paparare/DriveAgent}}

\end{abstract}

%%%%%%%%%%%%%%%%%%%%%%%%%%%%%%%%%%%%%%%%%%%%%%%%%%%%%%%%%%%%%%%%%%%%%%%%%%%%%%%%
\section{Introduction}
Promising progress has been made in autonomous driving (AD) in recent years; however, some challenging problems in AD have yet to be solved, especially under dynamic, multimodal environments, such as contextual understanding and interpretability~\cite{mckinsey2023autonomous}. Commonly adopted AD architectures, whether modular or end-to-end, often struggle to integrate insights across heterogeneous sensor modalities—such as cameras, LiDAR, IMU and GPS—especially in edge cases where visual information is ambiguous or missing~\cite{xie2024largemultimodalagentssurvey}. 

Similarly to AD problems, strong capabilities in reasoning across diverse domains have been demonstrated with large language models (LLM) and vision-language models (VLM) in the past few years~\cite{touvron2023llama,liu2023visual}. Nevertheless, a key challenge, how to apply LLM into multimodal sensor fusion in driving scenarios, remains underexplored~\cite{sun2024survey,zhang2024vision}. Thus, an opportunity to enhance autonomous decision-making is presented by incorporating LLM-driven reasoning into sensor-rich driving pipelines. 

\begin{figure}[!ht]
    \centering
    \includegraphics[width=\linewidth]{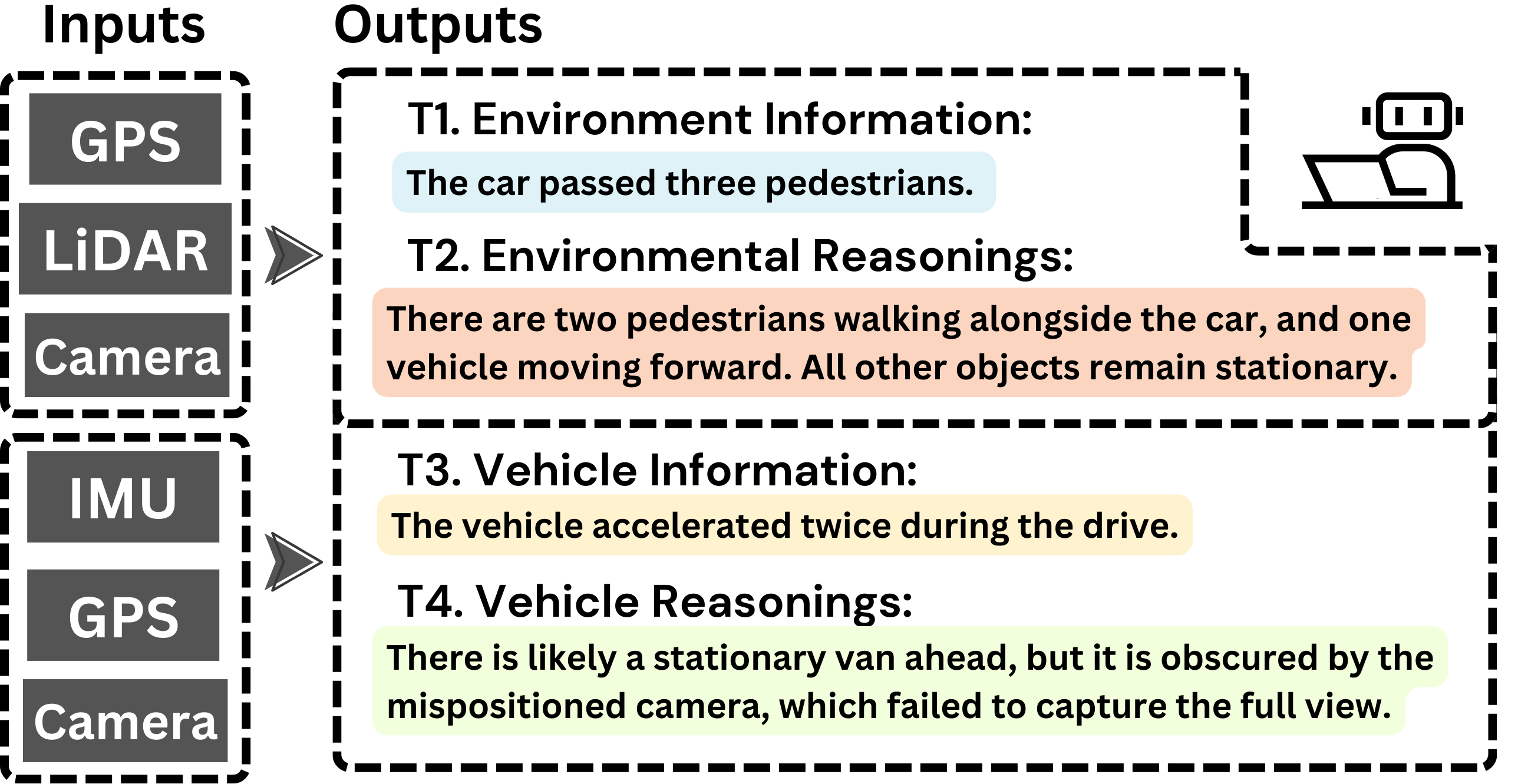}
    \caption{%
       Overview of the inputs and outputs for the proposed DriveAgent framework. DriveAgent takes multimodal sensor data as inputs, including camera, Lidar GPS, and IMU data. The input data are processed through four structured tasks (T1 to T4), supporting comprehensive reasoning tasks at both the environment level and the vehicle level as outputs.
    }
    \label{fig:overview}
\end{figure}

Recent research has begun exploring the integration of LLMs into autonomous driving tasks. For example, DriveLM~\cite{Sima2024DriveLM} proposed structured reasoning around visual input, and V2V-LLM~\cite{Chiu2025V2VLLM} advanced cooperative multimodal communication between vehicles. Additionally, frameworks such as GenFollower~\cite{lan2024genfollower} and LMDrive~\cite{shao2024lmdrive} have emphasized instruction-following and human-like behavior modeling. Similarly, prompting techniques have also advanced LLMs by improving reasoning and problem-solving. LaMPilot~\cite{ma2024lampilot} and KoMA~\cite{Jiang2024KoMA} both leveraged language-based prompting agents for decision-making, while TreeOT~\cite{Yao2023TreeOT} and ReActSR~\cite{Yao2023ReActSR} both proposed a similar method prompting LLMs to explore multiple reasoning paths, enhancing deliberate problem-solving, reasoning, and acting. However, current approaches concentrate narrowly on closed-loop planning, or single-task prompting, and use basic reasoning reliant only on relative object positions for visual understanding. As a result, they struggle to generalize to varied driving scenarios where visual sensors are unreliable—for instance, when cameras are misaligned or during hazardous driving conditions.

% Recent research has begun exploring the integration of LLMs into autonomous driving tasks. For example, LaMPilot~\cite{ma2024lampilot} and KoMA~\cite{Jiang2024KoMA} have leveraged language-based agents primarily for decision-making benchmarks. DriveLM~\cite{Sima2024DriveLM} proposed structured reasoning around visual input, and V2V-LLM~\cite{Chiu2025V2VLLM} advanced cooperative multimodal communication between vehicles. Additionally, frameworks such as GenFollower~\cite{lan2024genfollower} and LMDrive~\cite{shao2024lmdrive} have emphasized instruction-following and human-like behavior modeling. Nevertheless, prompting techniques have advanced LLMs by improving reasoning and problem-solving. ExpertPrompting~\cite{Xu2023ExpertPrompting} enhances response quality by synthesizing expert identities via in-context learning. TreeOT~\cite{Yao2023TreeOT} and ReActSR~\cite{Yao2023ReActSR} both proposed a similar method allowing LLMs to explore multiple reasoning paths, enhancing deliberate problem-solving, reasoning, and acting. However, most existing approaches concentrate narrowly on closed-loop planning or single-task prompting and rely on simple reasoning that uses only relative object positions for visual understanding. As a result, they struggle to generalize to varied driving scenarios where visual sensors are unreliable—for instance, when cameras are misaligned or during hazardous driving conditions.

Motivated by the aforementioned limitations, we introduce \textbf{DriveAgent}—a modular, LLM-driven multi-agent framework designed to reason over multimodal sensor streams in autonomous driving scenarios. DriveAgent integrates camera, LiDAR, GPS, and IMU data through a hierarchy of specialized agents that perform perception, reasoning, and decision-making tasks in a coordinated manner. Our framework leverages the structured compositionality of LLMs and domain-specific sensor processing modules to deliver clear, reliable responses across both typical and challenging driving situations. Unlike prior works that focus on end-to-end planning or vision-language alignment alone~\cite{xu2023drivegpt4,park2024vlaad}, a generalizable architecture is offered by DriveAgent to explain vehicle behavior, environmental dynamics, and causal events across multiple sensor types.

Fig.~\ref{fig:overview} illustrates our proposed study’s scope, showing how multimodal sensor inputs (e.g., camera, LiDAR, GPS, and IMU data) and text data support both vehicle-level and environmental-level tasks. Our contributions include:
\begin{enumerate}
    \item \textbf{Multi-Modal Agent System}: The proposed multi-modal agent system enables cohesive, end-to-end reasoning in complex driving contexts.
    
    \item \textbf{Vision-Language Model Fine-tune Strategy}: The proposed fine-tuned VLM enables abilities including object detection and traffic interpretation for the proposed system.
    
    \item \textbf{Self-Reasoning Benchmarks}: Autonomous driving performance is evaluated based on tasks such as data analysis, visual reasoning, and integrated environment understanding.
    
    \item \textbf{Three-Tier Driving Dataset}: The collected dataset represents standard, typical, and challenged AD scenarios, offering distinct challenges for comprehensive training and evaluation.
\end{enumerate}

\section{Methodology}
\label{section:methodology}

\begin{figure*}[htbp]
    \centering
    \includegraphics[width=0.75\linewidth]{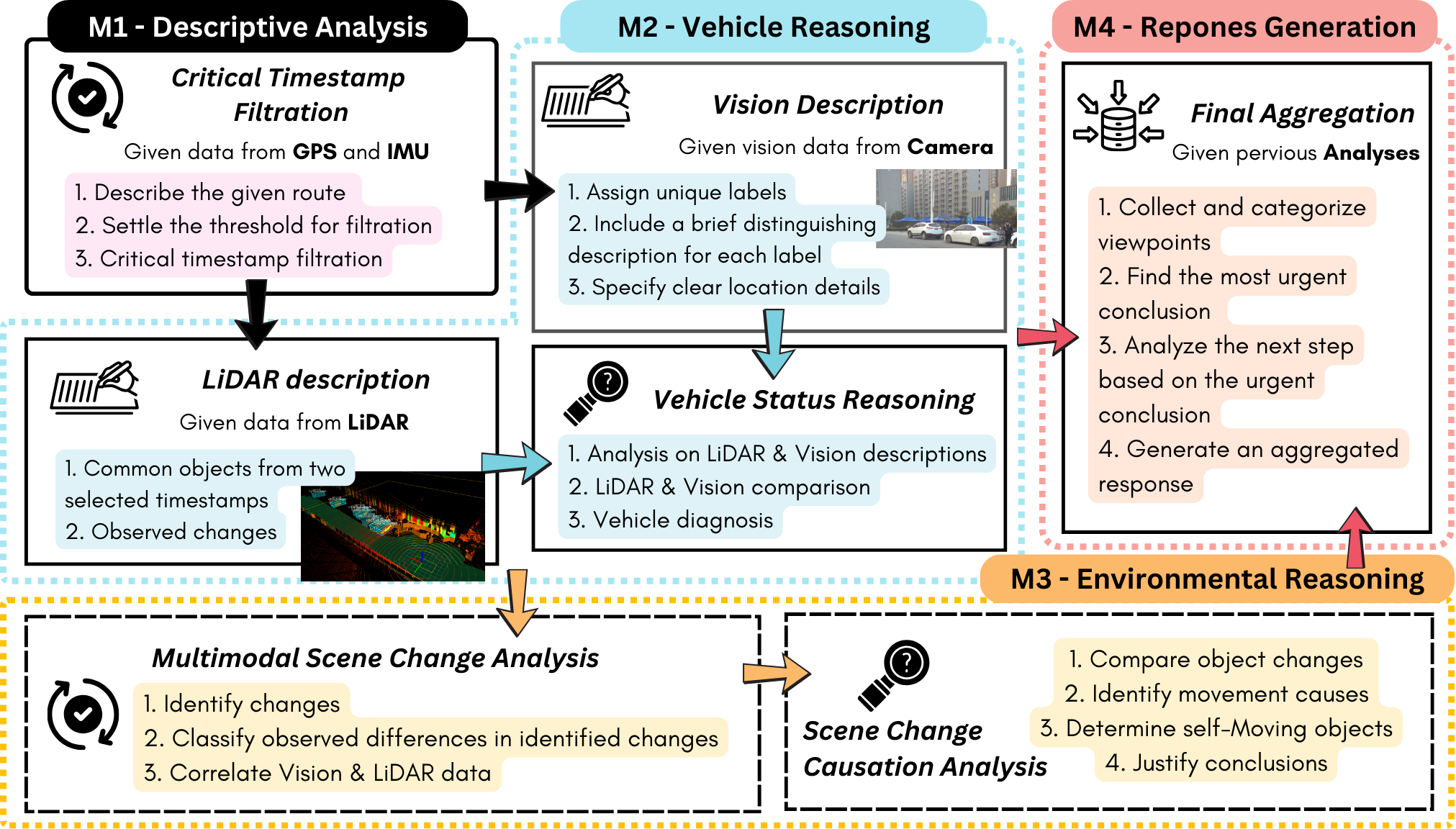}
    \caption{%
       An overview of the proposed architecture which is consisting of four modules (M1 to M4), where multimodal sensor inputs—camera, IMU, GPS and LiDAR—enable both environment-level tasks (e.g., information retrieval, environmental change detection, and reasoning) and vehicle-level tasks (e.g., vehicle status analysis, motion evaluation, and behavior pattern recognition). 
    }
    \label{fig:overview2}
\end{figure*}

Our approach addresses four key tasks through a structured reasoning process. Given an input instruction \( \mathcal{I} \), the module \( \mathcal{M} \) produces a response \( \mathcal{R} \) in adherence to the prompt. To facilitate driving analysis, we design four sequential modules as demonstrated in Fig.~\ref{fig:overview2}: \textbf{(1) Descriptive Analysis}, \textbf{(2) Vehicle Reasoning}, \textbf{(3) Environmental Reasoning}, and \textbf{(4) Response Generation}.

% \textcolor{red}{Jinglun Comments: to Xinmeng, I have make the whole section III more compact and updated some notations for disambiguation. Please update your figure 2 by the following points. 1. change Final Aggregation to Response Generation, in order to match with subsection title and algorithm title; 2. please add word: module in front of each color block, such as: Module 1. Descriptive Analysis; Module 2. Vehicle Reasoning...}

In the first phase, the system selects $n$ critical timestamps where significant events occur. We denote these timestamps and their triggering factors as $\{(T_i, F_i)\}_{i=0}^{n}$, where $T_i$ is the $i$-th timestamp and $F_i$ is the factor that prompted its selection. This set of time-factor pairs forms the basis for all subsequent analyses. The vehicle-reasoning phase consists of two independent sensor agents and one integration agent. The LiDAR agent $\mathcal{M}_L$ produces triplets $\{(T_i, F_i, L_i)\}_{i=0}^{n}$, where $L_i$ is the LiDAR-based description at time $T_i$. Similarly, the vision agent $\mathcal{M}_V$ produces $\{(T_i, F_i, V_i)\}_{i=0}^{n}$, with $V_i$ being the vision-based description at $T_i$. An aggregator agent $\mathcal{M}_D$ then compares each LiDAR description $L_i$ with the corresponding vision description $V_i$ to diagnose potential vehicle anomalies $D_i$. In parallel, an environmental reasoning agent uses $V_i$ and $L_i$ to analyze changes in the surrounding environment between consecutive timestamps. It identifies environment variations $E_{i+1}$ between times $T_i$ and $T_{i+1}$ (yielding changes $\{E_2, E_3, \dots, E_{n}\}$) and passes them to a causal analysis agent $\mathcal{M}_C$. The causal analysis agent uncovers the mechanisms behind each detected change and flags any objects requiring heightened caution as $C_i$. Finally, the response aggregation agent $\mathcal{M}_R$ consolidates the vehicle diagnostics $D_i$ from $\mathcal{M}_D$ and the caution flags $C_i$ from $\mathcal{M}_C$, and synthesizes them into a final response $\mathcal{R}_i$ for each critical timestamp $T_i$. Each $\mathcal{R}_i$ thus contains both the vehicle’s condition diagnosis ($D_i$ from the sensor comparison) and the relevant environmental and causal information ($C_i$ indicating any cautionary context).

\subsection{\textbf{Module 1: Descriptive Analysis}}

Determining which information is crucial for an accurate route description is a fundamental challenge in route analysis. We address this with a self-referential filtration system that automatically identifies critical timestamps based on the vehicle's motion. The filtration threshold is determined by an LLM agent analyzing prototypical route descriptions, from both real and simulated autonomous driving on predefined paths. A single agent handles route classification and threshold selection via this mechanism. 

% \textcolor{red}{Jinglun Comments: to Xinmeng, what does hat symbol represent in equation 1 and 2??.}

We categorize driving routes based on their speed $S$ and an urban complexity indicator $U$. Specifically, we define the function
\(
{\mathcal{R}}(S, U),
\)
which outputs both a route category $r_i$ and a corresponding threshold ${\theta}_i$. Formally (the double colon $::$ indicates this correspondence):
\[
{\mathcal{R}}(S, U) \in \{\,r_1 :: {\theta}_1,\; r_2 :: {\theta}_2,\; r_3 :: {\theta}_3\}, \tag{1}
\]
where $r_1$ represents high-speed, low-complexity routes, $r_2$ represents medium-speed, medium-complexity routes, and $r_3$ represents variable-speed, high-complexity routes. For each category $r_i$, ${\theta}_i$ is computed by an agent function $G$:
\[
{\theta}_i = G(S, U, r_i), \tag{2}
\]
which tailors standard kinematic baselines (angular velocity of $10^{\circ}/s$, linear acceleration of $8\,m/s{^2}$, and yaw rate of $10^{\circ}/s$) to the specific speed $S$ and urban complexity $U$. By monitoring these kinematic signals, such as turning, acceleration/braking, and orientation changes, the filtration agent efficiently pinpoints critical timestamps reflecting significant motion changes.

\subsection{\textbf{Module 2: Vehicle Reasoning}}
The Vehicle Reasoning module comprises three agents: one processing vision data, one processing LiDAR data, and an analyzer agent that synthesizes both to detect vehicle abnormalities. The designed reasoning pipeline is shown in Algorithm~\ref{alg:vehicle-symbolic-loop}.

\subsubsection{Vision Descriptor}
The vision agent first assigns unique labels to all detectable objects in the camera view, where each object gets an index $i$. It then examines two consecutive frames at times $t$ and $t+1$, recording the position of each object $i$ as $p_i(t)$ and $p_i(t+1)$. By comparing these positions, the agent measures how each object moved between the timestamps and can also derive an overall average movement across all objects, which are denoted as $p_i(t) \sim p_i(t+1)$ for each $i$. This relative position-change analysis identifies which objects have moved and by how much, providing a per-object motion summary between $t$ and $t+1$.

\begin{algorithm}[h!]
\caption{Vehicle Reasoning}
\label{alg:vehicle-symbolic-loop}
\begin{algorithmic}[1]

\REQUIRE $\{p_i(t)\}$ for vision data at $t=1,\dots,T$, $\{p_i(t)\}$ for LiDAR data at $t=1,\dots,T$, $\mathbf{L}_i(t), \mathbf{C}_i(t)$ for LiDAR/camera positions, $R$: distance threshold (e.g., $100$)

\FORALL{$t \in \{1,\dots,T-1\}$}

  \FORALL{$i$}
    \STATE $p_i(t)\,\sim\,p_i(t+1)$
  \ENDFOR

  \FORALL{$i$}
    \STATE $\Delta p_i = p_i(t+1) - p_i(t)$
  \ENDFOR

  \STATE $\Omega \gets \bigl\{\,i \mid \|\mathbf{L}_i(t)\|\le R\bigr\}$
  \FORALL{$i \in \Omega$}
    \STATE $\Delta_i(t) = \|\mathbf{L}_i(t) - \mathbf{C}_i(t)\|$
  \ENDFOR

\ENDFOR

\RETURN $\{\Delta p_i,\;\Omega,\;\Delta_i(t)\}$
\end{algorithmic}
\end{algorithm}

\subsubsection{LiDAR Descriptor}

The LiDAR agent begins with the set of object labels (as identified in the LiDAR point cloud) and their positions relative to the vehicle. If multiple objects initially share the same label $L_i$, the agent disambiguates them by spatial separation or other distinctive features to ensure each object $i$ is uniquely identified. It then considers two successive timestamps $t$ and $t+1$ and obtains object $i$’s positions $p_i(t)$ and $p_i(t+1)$ from the LiDAR data. The change in position is computed as:

\[
\Delta p_i = p_i(t+1) - p_i(t) \tag{3}
\]

% The LiDAR descriptor agent begins with a list of object labels, each corresponding 
% to an object's position relative to the vehicle. If multiple objects share the 
% same label \(L_i\), the agent distinguishes among them by examining their positions 
% or other discriminative features. Once each object is uniquely identified, the agent 
% considers two timestamps, \(t\) and \(t + 1\). Let \(p_i(t)\) and \(p_i(t+1)\) denote 
% the positions of object \(i\) (with label \(L_i\)) at these respective timestamps. 
% The position difference is then:
% \[
% \Delta p_i = p_i(t+1) - p_i(t),
% \]
% which can be used to measure how the object moves over time.

\subsubsection{Vehicle Status Reasoning}

The analyzer agent takes the outputs of both the vision and LiDAR descriptors to diagnose the vehicle’s status and sensor integrity. As a first step, it filters out any objects beyond a 100\,m range in the LiDAR data. Formally, it limits attention to the set $\Omega = \left\{ i \mid \| \mathbf{L}_i(t) \| \le 100 \right\}$, where $\mathbf{L}_i(t)$ is the LiDAR-derived position of object $i$ at time $t$ (in meters). This focuses the analysis on nearby objects and also allows a preliminary check for LiDAR sensor issues (e.g., if no objects appear within range when expected, the LiDAR could be malfunctioning or noisy). For each object $i \in \Omega$, the agent then compares its LiDAR position to the corresponding camera-inferred position. Let $\mathbf{C}_i(t)$ be object $i$’s position as estimated from the camera at time $t$. We define a consistency measure between the two sensors as the Euclidean distance:

\[
\Delta_i(t) = \bigl\| \mathbf{L}_i(t) \sim \mathbf{C}_i(t) \bigr\| \tag{4}
\]

If $\Delta_i(t)$ is large for a particular object, it suggests a discrepancy between LiDAR and camera, potentially due to calibration error or sensing noise. The agent also monitors if many objects exhibit large $\Delta_i(t)$ values simultaneously, which would indicate a broader sensor misalignment or a camera issue (e.g., blurring or calibration drift affecting $C_i(t)$ for multiple objects). After these checks, the agent compiles an integrated status report diagnosing any detected issues with the LiDAR data, such as missing/ghost objects or range errors; and camera data, such as poor object localization. 

% First, evaluate the LiDAR data and filter out any objects located more than 100\,meters away. Formally, let
% \(
% \Omega = \left\{ i \mid \| \mathbf{L}_i(t) \| \le 100 \right\}
% \)
% be the set of objects within 100\,meters, where \(\mathbf{L}_i(t)\) denotes the LiDAR-derived position of object \(i\) at time \(t\). At this stage, also diagnose potential LiDAR issues, such as sensor malfunctions or noisy readings.
% For each object \(i \in \Omega\), use both LiDAR and camera data to compare its position and track changes over time. Let \(\mathbf{C}_i(t)\) denote the position of the same object as inferred from camera data. A simple measure of consistency between the two sensors is the positional difference:
% \[
% \Delta_i(t) = \bigl\| \mathbf{L}_i(t) \sim \mathbf{C}_i(t) \bigr\|.
% \]
% If \(\Delta_i(t)\) is large, it may suggest a calibration issue or another source of inconsistency between the LiDAR and camera readings.
% Next, diagnose possible camera problems, such as unclear vision (leading to unreliable \(\mathbf{C}_i(t)\)) or misalignment. If many \(\Delta_i(t)\) values are unexpectedly large, it could point to systematic camera issues.
% Finally, the agent provides an integrated analysis that includes diagnoses of potential issues with both LiDAR and camera data. This consolidated report can then inform other downstream modules (e.g., perception and planning) about data quality and sensor reliability.

\begin{algorithm}[h!]
\caption{Environmental Reasoning}
\label{alg:symbolic-environmental}
\begin{algorithmic}[1]

\REQUIRE $\mathcal{V}(t)$, $\mathcal{V}(t-1)$: sets of visual detections at times $t$ and $t-1$, $\mathcal{L}(t)$, $\mathcal{L}(t-1)$: sets of LiDAR measurements at times $t$ and $t-1$, $\mathbf{O}_i(t)$: position of object $i$ at time $t$, $\Delta t$: time interval used for change detection in object positions

\FORALL{$v_i(t) \in \mathcal{V}(t)$}
  \STATE $\,\Delta v_i(t) \;=\; v_i(t)\;-\;v_i(t-1)$
\ENDFOR

\FORALL{$\ell_j(t) \in \mathcal{L}(t)$}
  \STATE $\,\Delta \ell_j(t) \;=\; \ell_j(t)\;-\;\ell_j(t-1)$
\ENDFOR

\STATE $\,\Delta_{i,j}(t) \;=\;\|{v}_i(t)\;-\;\mathbf{\ell}_j(t)\|\,$

\FORALL{$\mathbf{O}_i(t)$}
  \STATE $\,\Delta \mathbf{O}_i(t)\;=\;\mathbf{O}_i(t)\;-\;\mathbf{O}_i(t - \Delta t)$
\ENDFOR

\end{algorithmic}
\end{algorithm}

\subsection{\textbf{Module 3: Environmental Reasoning}}

Environmental reasoning module consists of two coordinated agents: one focused on detecting and characterizing environmental changes, and another dedicated to analyzing the causes of those changes. Working together, these agents provide a comprehensive understanding of the factors driving each observed environmental change as shown in Algorithm~\ref{alg:symbolic-environmental}.

\subsubsection{Environmental Reasoning}
This agent identifies environmental changes by comparing current sensor readings with those from the previous timestamp. Let $\mathcal{V}(t) = \{v_1(t), v_2(t), \dots, v_m(t)\}$ and $\mathcal{L}(t) = \{\ell_1(t), \ell_2(t), \dots, \ell_n(t)\}$ denote the vision and LiDAR detections at time $t$, respectively. By analyzing differences between $\mathcal{V}(t)$ and $\mathcal{V(}t-1)$ as well as $\mathcal{L}(t)$ and $\mathcal{L}(t-1)$, the agent detects new, missing, or significantly moved objects. Detected changes are classified based on type (e.g., static vs. dynamic) and severity. For each change, the agent also evaluates cross-sensor consistency. Given an object seen by both sensors, let $v_i(t)$ and $\ell_j(t)$ denote the positions of the same object as perceived by the camera and LiDAR, respectively. The sensor agreement can be quantified by the Euclidean distance between these estimates:

\[
\Delta_{i,j}(t) = \|{v}_i(t) - \mathbf{\ell}_j(t)\| \tag{5}
\]

Note that a small $\Delta_{i,j}(t)$ indicates that the vision and LiDAR agree on the object’s position, whereas a large $\Delta_{i,j}(t)$ could signal sensor misalignment, calibration issues, or an actual abrupt environmental change that one sensor registers differently than the other.

\subsubsection{Causal Analysis}

This agent investigates the underlying causes of the changes identified above. It first retrieves the state of each relevant object from prior reasoning stages or raw sensor data, denoting the position (or state) of object $i$ at time $t$ as $\mathbf{O}_i(t)$. It then looks at how each object’s state has evolved over a longer interval $\Delta t$ by computing and flags any object with a significant change $\Delta \mathbf{O}_i(t)$ for deeper analysis:

\[
\Delta \mathbf{O}_i(t) = \mathbf{O}_i(t) - \mathbf{O}_i(t - \Delta t)  \tag{6}
\]
For each flagged change, the agent infers plausible causes by analyzing temporal patterns (e.g., sudden vs. gradual), environmental cues (e.g., wind or collisions), and surrounding context (e.g., nearby object motion). It classifies each change as either \emph{self-moving} (e.g., vehicles or pedestrians) or \emph{externally influenced} (e.g., displaced by force), using cues like mobility features and motion behavior. The agent then compiles a causal report summarizing the changes, inferred origins, and confidence levels, enabling informed downstream decision-making with interpretable reasoning.

\subsection{\textbf{Module 4: Response Generation}}

This module synthesizes outputs from previous agents to generate a prioritized response, each insight $a_i$ is paired with a category $c_i$ (e.g., safety, efficiency), forming the set $\mathcal{A} = \{(a_i, c_i)\}_{i=1}^{N}$. A scoring function $\Psi(a_i, c_i)$ evaluates urgency, and the highest-priority issue, $\hat{a}$, can be identified as:

\[
\hat{a} = \arg\max_{(a_i, c_i) \in \mathcal{A}} \Psi(a_i, c_i) \tag{7}
\]

The agent then selects the best response $\phi^*$ from a candidate set $\Phi(\hat{a}) = \{\phi_1, \dots, \phi_M\}$ by maximizing a utility function:

\[
\phi^* = \arg\max_{\phi_j \in \Phi(\hat{a})} \mathrm{Score}(\phi_j) \tag{8}
\]

The final response is:

\[
\mathcal{R} = \bigl(\hat{a}, \phi^*, \mathcal{A}^-\bigr) \tag{9}
\]

where $\mathcal{A}^- = \mathcal{A} \setminus \{\hat{a}\}$ denotes secondary insights. This structured output can gather the top-priority issue, proposed action, and remaining considerations to support transparent and interpretable decision-making.

% \begin{algorithm}[h!]
% \caption{Response Generation}
% \label{alg:final-response}
% \begin{algorithmic}[1]

% \REQUIRE $\mathcal{A} = \{(a_i, c_i)\}$

% \FORALL{$(a_i, c_i) \in \mathcal{A}$}
%   \STATE $a_i \mapsto c_i$
% \ENDFOR

% \FORALL{$(a_i, c_i) \in \mathcal{A}$}
%   \STATE $Psi(a_i,c_i) \gets \mathrm{urgency}$
% \ENDFOR

% \STATE $\hat{a} \gets \displaystyle \arg\max_{(a_i, c_i) \in \mathcal{A}} \Psi(a_i, c_i)$

% \STATE $\Phi(\hat{a}) \gets \{\phi_j\}$
% \STATE $\phi^* \gets \displaystyle \arg\max_{\phi_j \in \Phi(\hat{a})} \mathrm{Score}(\phi_j)$

% \STATE $\mathcal{A}^- \gets \mathcal{A} \setminus \{\hat{a}\}$
% \STATE $\mathcal{R} \gets (\hat{a}, \phi^*, \mathcal{A}^-)$

% \RETURN $\mathcal{R}$

% \end{algorithmic}
% \end{algorithm}

% fintuned-LLaMA
% Accuracy: 0.6500
% Precision: 0.8440
% Recall: 0.6802

\section{Experiments}

\subsection{Datasets}

% \textcolor{red}{Jinglun Comments: to Long: can we add a reference for this algorithm: Xinda vehicle's LiDAR-based perception algorithm.}

% \textcolor{red}{Jinglun Comments: to Xinmeng, I updated the Section IV's structure for a clearer demonstration.}

% \textcolor{red}{Jinglun Comments: to Everyone, please update your figure/table reference properly in your writings, DO NOT use hard reference, please using $ref$ syntex instead.}

% \textcolor{red}{Jinglun Comments: to Wuqi and Long, what are Dataset question-answerings?}
\begin{table*}[ht]
\centering
\begin{tabularx}{\textwidth}{l c c >{\centering\arraybackslash}X}
\toprule
\textbf{Sensor} & \textbf{Model} & \textbf{Frequency} & \textbf{ Specifications}\\
\midrule
LiDAR & Robosense Ruby-128 & 10\,Hz & 128-beam, Maximum detection range: 230\,m, Vertical FOV: -25° to +15°, Horizontal FOV: 360° \\
\midrule
Camera & Basler acA2440-75uc & 10\,Hz & Default resolution: 2448\(\times\)2048 \\
\midrule
GNSS/INS & Xsens MTI-680G & 4\,Hz (GNSS), 400\,Hz (IMU/RTK) & 0.2° roll/pitch, 0.5° Yaw/Heading, cm-level position accuracy, Internal u-blox ZED F9 RTK-enabled GNSS receiver \\
\bottomrule
\end{tabularx}
\caption{Sensor Specification in Chang'an University Xinda Autonomous Vehicle.}
\label{tab:sensor_specification}
\end{table*}

\begin{figure}[htbp]
    \centering
    \includegraphics[width=0.9\linewidth]{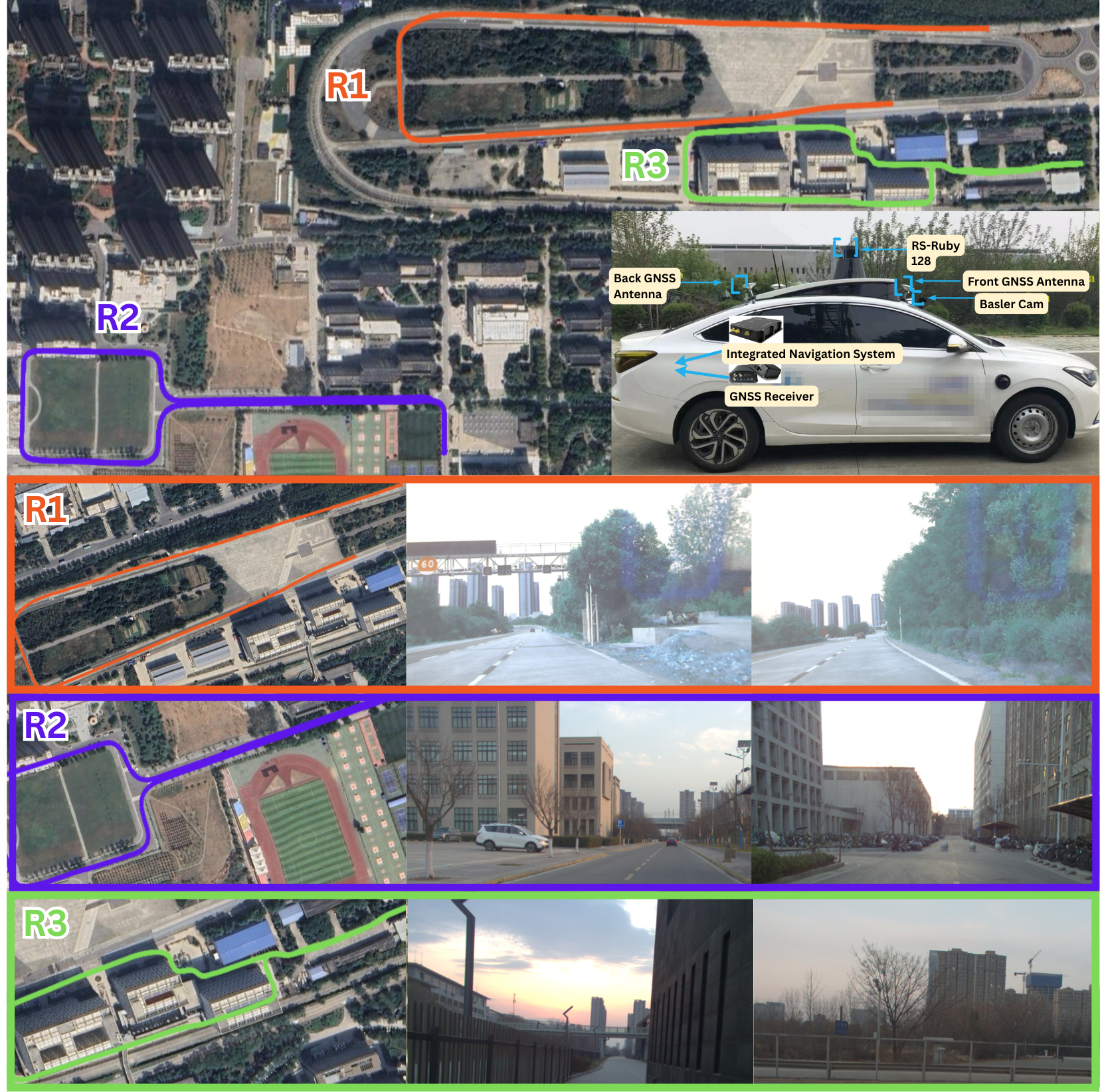}
    \caption{%
            Data collection vehicle sensor configuration and satellite images of recorded driving routes. There are three routes being recorded in total at Chang'an University, Xi'an, China. Route 1 (R1) is shown in red trajectory, Route 2 (R2) is shown in purple trajectory, while Route 3 (R3) is shown in green trajectory.
    }
    \label{fig:Equipment&datasets}
\end{figure}

Due to the lack of public datasets for evaluating an agent’s understanding of driving environments, we introduce a new dataset collected from an autonomous vehicle in real-world scenarios~\cite{wang2024modular}. As shown in Fig.~\ref{fig:Equipment&datasets}, the vehicle was equipped with multiple sensors and a navigation system \footnote{Datasets available at \url{https://huggingface.co/datasets/Parechan/driveagent}}. All sensor data were time-synchronized for consistent multi-modal observations. Sensor specifications are provided in Table~\ref{tab:sensor_specification}.

\begin{table}[ht]
\centering
\small
\renewcommand{\arraystretch}{1.1}
\setlength{\tabcolsep}{5pt}
\begin{tabularx}{\columnwidth}{lccc}
\toprule
\textbf{Attribute} & \textbf{R1} & \textbf{R2} & \textbf{R3} \\
\midrule
Length (m) & 1277.76 & 969.19 & 1125.91 \\
Max Speed (m/s) & 13.90 & 11.40 & 12.09 \\
Average Speed (m/s) & 7.30 & 4.17 & 4.29 \\
Environment Dynamic Level  & Small &  Large & Medium  \\
Roadside Obstructions & \ding{55} & \ding{55} & \ding{51} \\
Righ \& Left-side Camera Views & \ding{55} & \ding{51} & \ding{51} \\
\bottomrule
\end{tabularx}
\caption{Detailed attributes for routes R1, R2, and R3.}
\label{tab:dataset_info}
\end{table}

Moreover, as summarized in Table~\ref{tab:dataset_info}, our dataset covers three distinct driving routes: R1, R2, and R3. R1 spans $1277.76$ meters and was recorded in a controlled environment, serving as the baseline scenario. The ego vehicle reached a maximum speed of $13.90m/s$ with an average speed of $7.30m/s$, and only forward-facing images were captured. The environment dynamic level for R1 is qualitatively described as Small, reflecting relatively simple traffic conditions. R2, measuring 969.19 meters in length, features a loop around an urban square and is qualitatively described as having a Large environment dynamic level, indicating a more complex and active driving environment. The maximum and average speeds along R2 were $11.40m/s$ and $4.17m/s$ respectively. Compared to R1, R2 includes right and left-side camera views, providing a broader field of view. R3, at 1125.91 meters, introduces additional environmental complexity, with roadside obstructions and is qualitatively described as having a Medium environment dynamic level, indicating moderately active traffic with added structural challenges. The maximum speed recorded was $12.09m/s$, with an average speed of $4.29m/s$. Similar to R2, R3 captures views from the right, left, and front cameras. 

% Moreover, as summarized in Table~\ref{tab:dataset_info}, our dataset covers three distinct driving routes: R1, R2, and R3. R1 spans $1277.76$ meters and was recorded in a controlled environment, serving as the baseline scenario. The ego vehicle reached a maximum speed of $13.90m/s$ with an average speed of $7.30m/s$, and only forward-facing images were captured. The environment dynamic level for R1 is $2.00$ on a scale from $0.00$ to $5.00$, reflecting relatively simple traffic conditions. R2, measuring $969.19$ meters in length, features a loop around an urban square where the environment dynamic level is higher at $4.00$, indicating a more complex and active driving environment. The maximum and average speeds along R2 were $11.40m/s$ and $4.17m/s$ respectively. Compared to R1, R2 includes right and left-side camera views, providing a broader field of view. R3, at $1125.91$ meters, introduces additional environmental complexity, with roadside obstructions and $3.00$ as the environment dynamic level, indicating moderately active traffic with added structural challenges. The maximum speed recorded was $12.09m/s$, with an average speed of $4.29m/s$. Similar to R2, R3 captures views from the right, left, and front cameras. 

In addition, an enhanced detection method, combined with PointPillars architecture~\cite{lang2019pointpillars} and a clustering strategy, was used to perform real-time perception on LiDAR observation results and detect objects.

% \begin{table*}[ht]
% \centering
% \renewcommand{\arraystretch}{1.1} % Reduce row height, default is 1.2, smaller = tighter
% \setlength{\tabcolsep}{4pt} % Reduce horizontal space between columns
% \begin{tabularx}{\textwidth}{l c c c c c}
% \toprule
% \textbf{Routes} & \textbf{Length (m)} & \textbf{Max Speed (m/s)}
% & \textbf{Average Speed (m/s)} & \textbf{Number of Dynamic Objective} & \textbf{Roadside Obstructions}\\
% \midrule
% R1 & 1277.76 & 13.90 & 7.30 & 2 & \ding{55} \\
% R2 &  969.19 & 11.40 & 4.17 & 4& \ding{55}\\
% R3 & 1125.91 & 12.09 & 4.29 & 3& \ding{51}\\
% \bottomrule
% \end{tabularx}
% \caption{Detailed information of the collected datasets on routes R1, R2, and R3.}
% \label{tab:dataset_info}
% \end{table*}

% number of timestamp * route number ==> dataset size

\subsection{Task and Evaluation Metrics}
\label{section:tasks}
% \textcolor{red}{Jinglun Comments: to Xinmeng, let's introduce evaluation metrics first before going into details.}

We define three primary tasks: (1) \textbf{Object and Category Detection}, (2) \textbf{Vehicle Reasoning} (LiDAR and visual understanding), and (3) \textbf{Environmental Reasoning}. Each task is validated by its contribution to scene understanding, decision-making, and system robustness, with results discussed in Section~\ref{section:results}.

For object identification task, we consider seven key categories: \textit{four-wheel vehicles} (the principal motorized participants on roads), \textit{non-four-wheel vehicles} (e.g., bicycles and scooters, which often pose higher risk due to less coverage), \textit{pedestrians} (vulnerable road users who commonly receive priority), \textit{signs} (official traffic instructions and regulations), \textit{fixed installations} (permanent structures, barriers, or buildings), \textit{plants} (vegetation that may obscure visibility or mark boundaries), and \textit{monitors} (electronic displays or cameras supporting traffic supervision). This task is trained on datasets {R2} and {R3} and evaluated on {R1}, using precision, recall, and F1 as metrics; its importance lies in ensuring the accurate classification of objects critical for traffic safety.

The vehicle-reasoning task include two tasks: a LiDAR understanding task, evaluated by comparing the model’s output with ground-truth labels in R2, and a vision-based reasoning task, assessed on R2 and R3, where misaligned camera views serve as distractors. These evaluations measure real improvements in perception accuracy and prevent false gains from random guessing. Finally, the environmental reasoning task tests the system’s ability to tell apart stationary objects from independently moving ones (like pedestrians), with improvements validated through better situational awareness, collision avoidance, and safer navigation in dynamic traffic.

\subsection{Baseline Approaches}

% \textcolor{red}{Jinglun Comments: to Xinmeng, let's also introduce baseline methods briefly. If there are not much to write, you can combine the evaluation metrics and baseline apprroaches together.}

For \textit{task 1}, we benchmark five leading vision-language models including LLaMA-3.2-Vision-Instruct~\cite{meta2024llama3.2}, GPT-4o-mini~\cite{openai2024gpt4o}, Pixtra-large~\cite{pixtra2024pixtra}, GPT-4o~\cite{openai2024gpt4o}, and Claude-3.7-Sonnet~\cite{anthropic2024claude}—selected for their strong performance, diverse architectures, and proven effectiveness on vision tasks.

For \textit{tasks 2 \& 3}, we adopt three baseline methods: Zero-Shot~\cite{brown2020language}, CoT~\cite{wei2022chain}, and CoT + Self-Refine~\cite{madaan2023selfrefine}. Zero-Shot tests direct inference ability, CoT adds step-by-step reasoning, and CoT + Self-Refine further improves reasoning through iterative refinement.

\begin{figure*}[htbp]
    \centering
    \includegraphics[width=0.9\linewidth]{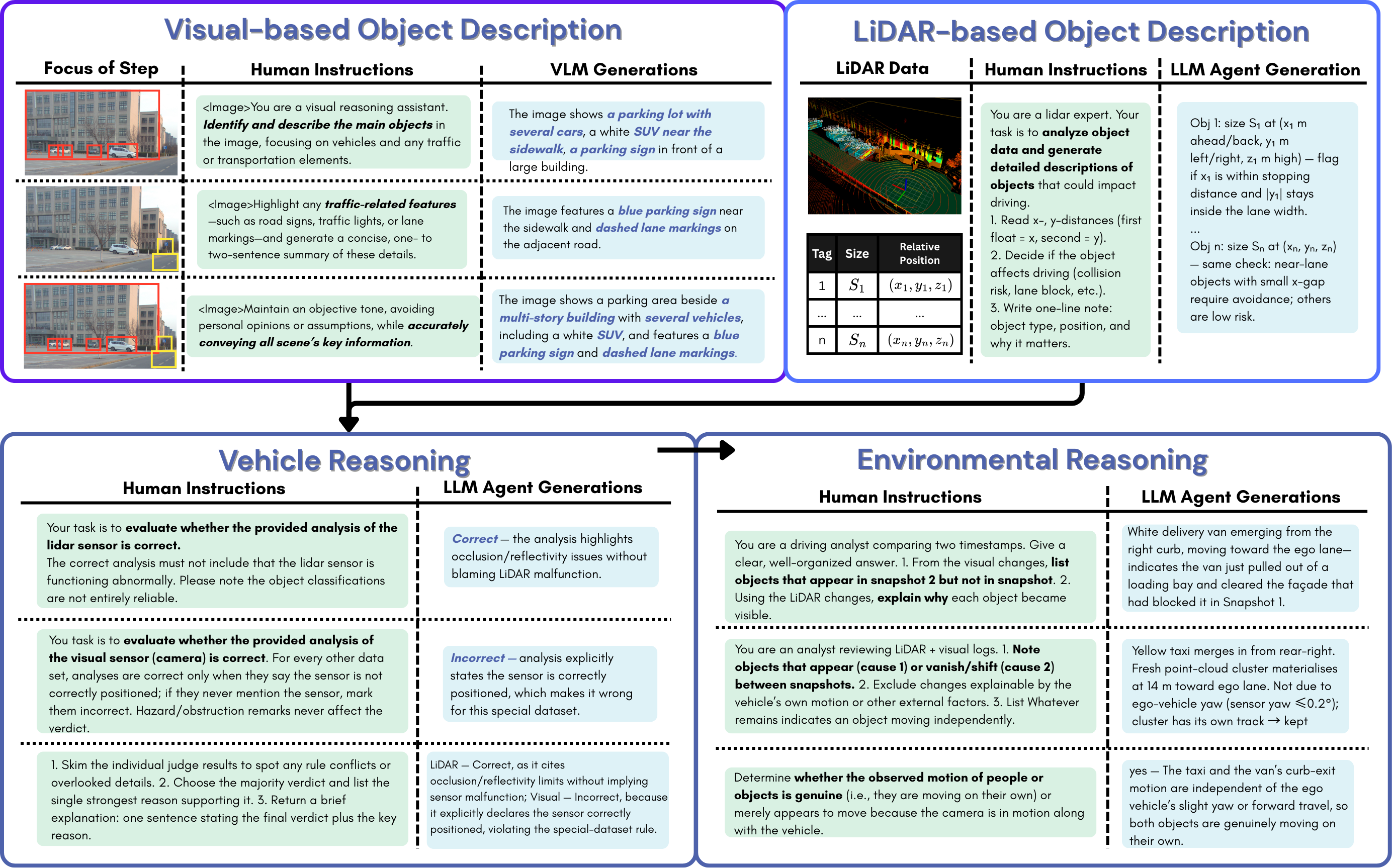}
    \caption{%
            Overview of the multimodal reasoning pipeline used for driving scene understanding. Visual descriptions are generated from camera images, focusing on identifying traffic-related objects and maintaining objective scene summaries. LiDAR-based descriptions analyze object sizes and relative positions to assess driving risk. In the reasoning stages, LLM agents evaluate the correctness of sensor-based analyses (vehicle reasoning) and identify environmental changes over time (environmental reasoning). Human instructions and corresponding LLM generations are provided for each step, supporting robust, explainable autonomous driving assessments.}

    \label{fig:example}
\end{figure*}

\subsection{Reasoning Instructions}

% \textcolor{red}{Jinglun Comments: to Xinmeng, can you update Table~\ref{tab:vlm_instructions} to a figure style like this example~\ref{fig:example}? I think this style is more vivid than a plain table :)}

Fig.~\ref{fig:example} outlines structured annotation guidelines that define the expected format and content of a high-quality response. These guidelines emphasize three key aspects: (1) correctly identifying vehicles and other dynamic traffic elements (e.g., bicycles, buses), (2) highlighting relevant static road infrastructure such as lane markings, traffic signs, and signals, and (3) ensuring that descriptions are objective, concise, and free from subjective or irrelevant content.

To assess the quality of the model’s outputs, we compare the generated descriptions against reference descriptions derived from these guidelines. The evaluation focuses on both content accuracy and coverage of key visual categories. Specifically, we extract five scene components from each output: Trees, Buildings, Vehicles, Pedestrians, and Signs. These categories are selected due to their relevance to road-scene understanding and their prevalence in standard autonomous driving datasets.

\textbf{Reasoning Setup:}
Reasoning experiments follow the multi-phase reasoning methodology described in Section~\ref{section:methodology}, the proposed DriveAgent is deployed to complete four sequential modules: Descriptive Analysis, Vehicle Reasoning, Environmental Reasoning, and Response Generation. For each phase, DriveAgent generates a response based on the intermediate input from the previous step, resulting in a total of four stepwise generations per input case. Evaluation is conducted at two critical points: (1) assessing the accuracy of the agent’s vehicle diagnostic reasoning, and (2) evaluating the accuracy of its environmental and causal reasoning.

\textbf{VLM Implementation Details:}
The VLM model in DriveAgent utilizes the LLaMA-3.2-Vision model (11B Parameter) as the foundation, where the vision tower is a pre-trained LLaMA vision encoder and the language model is LLaMA-3.2 (11B). A learnable linear image projection layer is inserted between the vision encoder and the LLM to align visual features with the LLM's input space. We fine-tune both the vision encoder and the LLM using Low-Rank Adaptation (LoRA)~\cite{hu2022lora}, while training the projection layer from scratch (i.e., no LoRA applied). All experiments are conducted on a server equipped with an NVIDIA H100 GPU. The model is optimized using the AdamW optimizer with an initial learning rate of $2\times10^{-4}$ and a batch size of 2. We employ a cosine learning rate decay schedule with a warm-up ratio of $0.03$. Training is performed for $10$ epoch using instruction-style supervision introduced in the VLM Instructions section, where each training sample is formatted as an instruction-response pair that includes special \textless \text{Image}\textgreater tokens to denote visual inputs. In practice, we construct structured JSON-formatted prompts containing both textual instructions and placeholder tokens for images, and fine-tune the model with supervised instruction tuning on these multimodal prompts.

\section{Results and Analysis}
\label{section:results}
\subsection{Object and Category Detection Performance}

In this subsection, we first evaluate the task 1 introduced at Section~\ref{section:tasks}. Table~\ref{tab:results} illustrates the substantial performance gains achieved when training with structured annotation guidelines aimed at better and more accurate object identification. The baseline LLaMA-3.2-Vision-Instruct model achieves moderate performance (Precision = $64.33\%$, Recall = $35.26\%$, F1-score = $45.55$). However, once annotation guidelines are systematically applied, the VLM model in DriveAgent exhibits a significant leap in all key metrics—reaching a precision of $89.96\%$ and an F1-score of $71.62$, outperforming other models in the table.

\begin{table}[ht]
\centering
\begin{tabular}{lccc}
\toprule
\textbf{Model} & \textbf{Precision} & \textbf{Recall} & \textbf{F1-score} \\
\midrule
LLaMA-3.2-Vision-Instruct & 64.33 & 35.26 & 45.55 \\
GPT-4o-mini & 64.40 & 51.60 & 57.30 \\
pixtra-large & 76.12 & 54.63 & 60.86 \\
GPT-4o  & 80.98 & 59.98 & 64.96 \\
claude-3.7-sonnet & 83.83 & \textcolor{red}{69.73} & 68.80 \\
\midrule
DriveAgent  & \textcolor{red}{89.96} & 63.52 & \textcolor{red}{71.62} \\
\bottomrule
\end{tabular}
\caption{Precision (\%), Recall (\%), and F1-score for each baseline methods and the proposed model on Object and Category Detection task. Red color highlights the best performance.}
\label{tab:results}
\end{table}

\begin{figure}[htbp]
    \centering
    \includegraphics[width=0.95\linewidth]{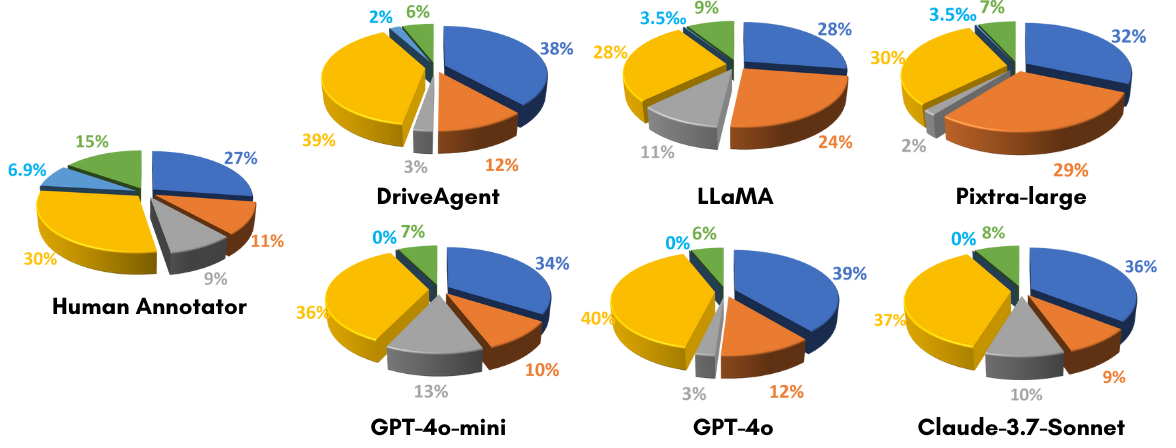}
    \caption{Distribution of object categories in the human-annotated ground truth versus each model’s predictions. Colour key: \textcolor{deepblue}{deep blue = fixed installations}, \textcolor{orange}{orange = four-wheel vehicles}, \textcolor{deepgray}{grey = non-four-wheel vehicles}, \textcolor{yellow!90!black}{yellow = plants}, and \textcolor{cyan}{light blue = monitors}.
    }
    \label{category}
\end{figure}

Fig.~\ref{category} shows that compared to the human annotator, DriveAgent is the only model that consistently detects monitors, while the other baselines mostly miss them, which is because overhead monitors are less conspicuous than ground-level objects. This improvement underscores the importance of precise, consistent labeling for training object detection systems. By removing ambiguity and ensuring uniform criteria for bounding boxes and class labels, the new annotations allow the model to learn object boundaries and distinctions more effectively. Consequently, DriveAgent demonstrates superior accuracy in localizing and identifying objects, validating that high-quality, structured annotation practices are crucial to achieving robust object identification performance.

\subsection{Reasoning Performance}
\textbf{Vehicle Reasoning:} We first We first evaluate the Vehicle Reasoning task (LiDAR and Vision) introduced in in Section~\ref{section:tasks}, as shown in Table~\ref{tab:reasoning_accuracies_categorized}.

For LiDAR reasoning, the Zero-Shot approach achieves moderate accuracy ($47.50\%$–$65.05\%$) across routes, establishing a baseline for detecting sensor misplacement. CoT alone leads to substantial performance drops, suggesting basic sequential reasoning struggles with subtle errors. Adding Self Refine significantly improves accuracy, reaching $72.63\%$ on R2 and $63.89\%$ on R2-right. DriveAgent, however, achieves strong and stable performance, particularly on R2-left ($69.90\%$), demonstrating reliable LiDAR misplacement detection.

For Vision reasoning, detecting misaligned cameras is even more challenging. Zero-Shot and CoT show very low accuracies on left and right views. In contrast, DriveAgent achieves notable gains, including $96.84\%$ accuracy on R2, and clear improvements across left and right variants ($58.25\%$ and $71.30\%$), confirming that modality-specific tuning is crucial for visual sensor reasoning.

\begin{table}[ht]
\centering
\scriptsize
\begin{adjustbox}{max width=\linewidth}
\begin{tabular}{l|l|cccccc}
\toprule
\textbf{Tasks} & \textbf{Method} & \textbf{R2} & \textbf{R2-left} & \textbf{R2-right} & \textbf{R3} & \textbf{R3-left} & \textbf{R3-right} \\
\midrule
\multirow{4}{*}{\shortstack[l]{Vehicle\\Understanding\\(Lidar)}} 
& Zero-Shot           & 62.11  & 65.05  & 52.78  & 57.14  & 47.50  & \textcolor{red}{55.26} \\
& CoT                 & 15.79  & 22.33  & 15.74  & 18.10  & 20.00  & 13.16 \\
& CoT + Self Refine   & \textcolor{red}{72.63}  & 66.02  & \textcolor{red}{63.89}  & \textcolor{red}{64.76}  & 45.00  & 55.26 \\
& DriveAgent (ours)   & 65.26  & \textcolor{red}{69.90}  & 58.33  & 55.24  & \textcolor{red}{51.25}  & 50.00 \\
\midrule
\multirow{4}{*}{\shortstack[l]{Vehicle\\Reasoning\\(Vision)}} 
& Zero-Shot           & 70.53  & 2.91   & 0.93   & 82.86  & 6.25   & 3.95 \\
& CoT                 & 69.47  & 25.24  & 27.78  & 79.05  & 31.25  & 40.79 \\
& CoT + Self Refine   & 65.26  & 0.97   & 2.78   & 80.00  & 3.75   & 5.26 \\
& DriveAgent (ours)   & \textcolor{red}{96.84}  & \textcolor{red}{58.25}  & \textcolor{red}{71.30}  & \textcolor{red}{87.62}  & \textcolor{red}{68.75}  & \textcolor{red}{63.16} \\
\midrule
\multirow{4}{*}{\shortstack[l]{Environmental\\Reasoning}} 
& Zero-Shot           & 37.89  & -      & -      & 36.19  & -      & - \\
& CoT                 & 56.84  & -      & -      & 62.86  & -      & - \\
& CoT + Self Refine   & 43.16  & -      & -      & 56.19  & -      & - \\
& DriveAgent (ours)   & \textcolor{red}{58.95}  & -      & -      & \textcolor{red}{65.71}  & -      & - \\
\bottomrule
\end{tabular}
\end{adjustbox}
\caption{Reasoning accuracy (\%) is reported across modalities and regions, with each task presenting results from several prompting methods. The labels R-left* and R-right* denote the left- and right-side camera views of the same route; these views act as distractors for the vehicle-reasoning subtask. Red color highlights the best performance.}
\label{tab:reasoning_accuracies_categorized}
\end{table}

\textbf{Environmental Reasoning:} At last, we evaluate the task 3 introduced in Section~\ref{section:tasks}. The evaluation of environmental reasoning performance is based on the agent's ability to detect independently moving objects by comparing two selected timestamps. As shown in Table~\ref{tab:reasoning_accuracies_categorized}, the Zero-Shot performance is low ($37.89\%$ and $36.19\%$), indicating that without any additional reasoning cues the agent struggles with temporal object differentiation. The CoT method significantly improves performance, achieving accuracies of $56.84\%$ and $62.86\%$. However, the performance of CoT $+$ Self Refine strategy offers mixed results, where the performance drops to $43.16\%$ for one set and recovers partially to $56.19\%$ for the other, suggesting that the refinement process may not always synergize effectively with the inherent sequential reasoning of CoT in this task. Notably, our proposed DriveAgent model outperforms all baselines, obtaining the highest accuracies of $58.95\%$ and $65.71\%$ respectively. These results underscore the importance of a dedicated, well-tuned approach for integrating temporal and spatial reasoning, which is critical for accurately identifying independently moving objects in dynamic environments.

\section{Conclusion}

In this paper, we propose {DriveAgent}, a modular, LLM-guided multi-agent framework for structured reasoning in autonomous driving. By integrating multimodal sensor inputs—camera, LiDAR, GPS, and IMU—into a hierarchy of perception and reasoning agents, DriveAgent addresses a long-standing challenge in autonomous systems: robust interpretation of complex, sensor-rich driving environments.

We demonstrate how the system’s four core modules—descriptive filtration, vehicle-level diagnostics, environmental reasoning, and urgency-aware response generation—jointly enable high-resolution understanding of both vehicle behavior and environmental context. Experiments on real-world multi-sensor datasets shows that DriveAgent not only surpasses baseline prompting approaches in both accuracy and stability, but also offers significant advantages in interpretability and modular extensibility. Moreover, the proposed three-tier dataset, along with self-reasoning benchmarks and structured VLM fine-tuning pipeline can further contributed to the broader community. Last but not least, DriveAgent provides a path forward for generalizable, interpretable, and sensor-aware autonomy. The proposed method bridges foundational advances in language modeling with real-time demands of perception and control—laying the groundwork for future driving systems that are not only reactive, but also reflectively aware.

\bibliographystyle{IEEEtran}  
\bibliography{IEEEfull}

\end{document}